# Sentiment Analysis of Spanish Political Party Tweets Using Pre-trained Language Models

Authors:
**Chuqiao Song**
Institution: Fudan University, Yangpu District, Handan Road 220, Shanghai, China
Email: 21300120147@m.fudan.edu.cn
**Shunzhang Chen**
Institution: Beijing Institute of Technology, Haidian District, Zhongguancun Avenue 5, Beijing, China
Email: 1120212450@bit.edu.cn
**Xinyi Cai**
Institution: Beijing Institute of Technology, Haidian District, Zhongguancun Avenue 5, Beijing, China
Email: 1120212476@bit.edu.cn
**Hao Chen**
Institution: Fudan University, Yangpu District, Handan Road 220, Shanghai, China
Email: h_chen@fudan.edu.cn

**Abstract:** This study investigates sentiment patterns within Spanish political party communications on Twitter by employing BETO and RoBERTuito, two pre-trained language models optimized for Spanish text. With a dataset comprising tweets from major Spanish political parties—PSOE, PP, Vox, Podemos, and Ciudadanos—spanning 2019 to 2024, this research analyzes sentiment distributions and explores the relationship between sentiment and party ideology. Results reveal that both models consistently identify a predominant Neutral sentiment across parties, with significant variations in Negative and Positive sentiments that align with ideological distinctions. Vox exhibits higher levels of Negative sentiment, while PSOE demonstrates a relatively high Positive sentiment, supporting the hypothesis that emotional appeals in political messaging reflect ideological stances. This study highlights the utility of pre-trained models in analyzing non-English social media sentiment and underscores the implications of sentiment dynamics in shaping public discourse within a multi-party system.

# 1. Introduction

In the era of digital politics, social media has emerged as a potent platform where public opinion is actively shaped and reflected. Political entities globally harness these platforms, with Twitter playing a central role in real-time political discourse. For countries like Spain, where a spectrum of political ideologies coexists, understanding the sentiment behind political communications becomes crucial. Sentiment analysis, particularly on platforms like Twitter, serves as a powerful tool to decode public attitudes and the emotional undertones in political party communications (Cambria et al., 2013; Giachanou & Crestani, 2016). By leveraging sentiment analysis, researchers can quantify and interpret political sentiments, thereby offering insights into party strategies and public reactions.

In Spain's unique political landscape, where new and traditional parties like Podemos, PSOE, PP, Ciudadanos, and Vox engage vigorously on social media, analyzing sentiment can reveal the underlying strategies each employs. Recent advancements in pre-trained models tailored for the Spanish language, such as BETO and RoBERTuito, offer refined accuracy in detecting nuanced sentiments within Spanish tweets (Pérez et al., 2021). These models not only account for linguistic variations but also address challenges posed by regional dialects and context-specific slang, enhancing their relevance in political sentiment analysis.

This study explores the efficacy of BETO and RoBERTuito in analyzing Spanish political tweets to detect correlations between expressed sentiment and party ideologies. By doing so, it contributes to a growing field of research aiming to enhance the precision of sentiment analysis in non-English contexts and offers a quantitative basis for assessing ideological sentiment in Spain's political communications.

# 2. Literature review
## 2.1 Emotion in Political Parties

In contemporary democratic societies, deliberative democracy emphasizes the importance of public dialogue, where individuals not only represent their interests but also defend their positions and consider opposing perspectives. This ideal goes beyond simple majority rule, encouraging citizens to engage in discourse that transcends personal bias (Slothuus and Bisgaard, 2021). Political leaders play a pivotal role in this dynamic, as their messages can profoundly shape public attitudes and behaviors, especially in times of crisis. For instance, government communication has been shown to influence public compliance with health directives, underscoring the need to better understand the impact of political messaging on individual behavior during crises (Peitz et al., 2021).

Prior research consistently demonstrates that emotions like happiness, hope, fear, hate, and anger significantly impact decision-making across various domains, including risk-taking, consumer ethics, and indulgence (e.g., Garg et al., 2005; Garg and Lerner, 2013). In political communication, emotional appeal—described as an "extreme emotional ingredient"—is particularly powerful. Populist parties, for example, often utilize highly emotional rhetoric to resonate with the concerns and frustrations of ordinary citizens, fostering a sense of urgency and solidarity among supporters (Canovan, 1999). Emotional responses can deeply influence how citizens process information, form attitudes, and ultimately make political decisions (Brader, Valentino, & Suhay, 2008; Kühne & Schemer, 2015; Vasilopoulos et al., 2018; Vasilopoulou & Wagner, 2017).

Research also highlights a distinct emotional strategy in political messaging between governing and opposition parties. For instance, Fourie et al. (2022) found that during a political crisis, opposition parties in Africa relied on appeals of uncertainty, fear, and rage, while the ruling party focused on messages of hope and achievement. On another continent, during the COVID-19 pandemic in the Czech Republic, populist and opposition parties were found to predominantly use negative emotions on social media (Kluknavská & Macková, 2024). These qualitative studies illustrate a common pattern: opposition parties, especially populist ones, tend to leverage negative emotions like fear and anger to challenge the status quo, while ruling parties may emphasize stability and progress through positive emotions.

In recent years, advancements in large language models have enabled quantitative analyses of emotional appeals in political communication. For example, Widmann et al. (2022) utilized a neural network classifier to analyze political discourse during the COVID-19 pandemic, revealing that populist radical parties increased hope appeals while decreasing fear rhetoric—a stark contrast to government parties, which amplified fear and downplayed hope. Similarly, Govind et al. (2023) employed the RoBERTa model to examine the impact of conservative and liberal state leaders' emotional language on public behavior. Their findings suggest that conservative leaders' use of hate correlated with increased nonessential driving, while appeals to hope decreased it; intriguingly, liberal leaders exhibited an opposite pattern.

These studies underscore the strategic use of emotional appeals in political messaging, especially in populist and opposition contexts, where emotions like fear, anger, and hope are carefully deployed to influence public opinion and behavior. The qualitative and quantitative insights from these studies not only reveal the emotional dynamics at play in political discourse but also highlight the broader implications for public compliance, social cohesion, and political polarization. In sum, understanding the nuances of emotional appeals in political communication is essential for grasping the full impact of political messaging on democratic processes, particularly in times of crisis.

**2.2 Sentiment Analysis**
Sentiment Analysis (SA), also referred to as Opinion Mining (OM), is a critical area of Natural Language Processing that seeks to understand and analyze human opinions, stances, and attitudes conveyed through written or spoken language. By analyzing textual or vocal data, sentiment analysis technologies can infer an author's or speaker's emotional state, encompassing their attitudes and emotional leanings toward particular subjects or themes. Key approaches in sentiment analysis include dictionary- and statistics-based methods, machine learning, and deep learning techniques (Pang, Bo & Lilian Lee, 2008; Liu, Bing, 2012; Zhang, Lei & Wang et al., 2018). These methods leverage various technologies to autonomously identify and extract emotional attributes and insights from text, enhancing sentiment analysis's precision and adaptability across diverse applications.

Sentiment analysis is widely used across fields such as social media analysis, customer service feedback, product evaluation, and public opinion monitoring, providing essential insights for informed decision-making and business strategy (Cambria, Erik & Björn Schuller et al., 2013; Yue, Zhang & Wang et al., 2019; Giachanou, Anastasia, & Fabio Crestani, 2016). According to Global WebIndex, as of January 2024, about 62.3% of the global population was active on social media, with an average daily use of 2 hours and 23 minutes. Given the wealth of data on social platforms, applying sentiment analysis technologies to social media not only yields extensive data but is increasingly necessary.

## 2.2 Sentiment Analysis on Social Media

Sentiment analysis has become widely utilized on social media, particularly on platforms like Twitter, where analyzing tweet sentiment enables researchers to accurately capture public attitudes toward products, services, and social issues—an asset in both business and politics. While tweets often include elements like URLs, emojis, and hashtags that add complexity to sentiment analysis, this application shows substantial promise in business, marketing, and political campaigns by equipping decision-makers with tools to better understand public opinion and shape strategic responses (Gupta & Kumar, 2023). Through in-depth customer sentiment analysis, companies can detect trends, patterns, and preferences, facilitating informed decision-making and improvements in customer service, product development, and marketing approaches. In politics, sentiment analysis is commonly used to predict election results, gauging public sentiment from social media to support campaign strategy (Chauhan et al., 2020). Specifically, it is integral to measuring public opinion for crafting campaign strategies (Ramteke, Shah & Godhia, 2016). Park et al. (2021) examined the temporal dynamics of emotional appeals in Russian election campaigns during the 2016 election, analyzing Twitter and Facebook data using word-level sentiment analysis and natural language processing (NLP) techniques. Their study found that parties used distinct emotional approaches for different platform audiences. Somula (2016) also analyzed public sentiment on platforms like Facebook and Twitter, classifying 2016 U.S. election tweets into positive, neutral, and negative categories to evaluate public sentiment toward the candidates.

Social media sentiment analysis predominantly involves sentence-level sentiment analysis (SLSA), which assesses individual sentences to determine whether they convey positive, negative, or neutral opinions. This level of analysis is closely connected to subjectivity classification, with Bongirwar et al. (2015) explaining that subjectivity classification distinguishes sentences with factual content (objective sentences) from those expressing personal views or opinions (subjective sentences), offering a clearer understanding of sentiment in context. Early SLSA research, such as by Tripathy et al. (2016), relied on manually designed text features (like lexicons and n-grams) for machine learning algorithms to classify text data. Advances in deep neural networks led Wang J et al. (2016) and Tang et al. (2015) to develop neural network-based methods (e.g., CNN, LSTM) for sentence-level sentiment classification. The introduction of attention mechanisms brought new methods to this task; for instance, Li et al. (2018) showed that combining attention mechanisms with neural networks improves the extraction of fine-grained emotional features within sentences.

To further enhance polarity detection, Su et al. (2023) noted that researchers are integrating traditional linguistic features into pre-trained language models to better tailor them for downstream tasks, as seen in Liu et al. (2019). Pre-trained models rely on the Transformer architecture, which employs self-attention to capture long-range dependencies within input sequences.

**2.3 Current Landscape of Sentiment Analysis in Spanish**
Currently, most sentiment analysis models are primarily designed for English, though research in languages like Chinese, Portuguese, and Korean exists. Despite this, English-based models remain dominant. Sentiment analysis in Spanish presents unique challenges. Although Spanish is one of the most widely spoken languages globally—with over 500 million speakers, surpassed only by English, Chinese, and Hindi, and being the second most-used language on Twitter (Fernández Vítores, 2020)—it has not yet received proportional research attention. Some Spanish sentiment lexicons are translated from English sources, such as SOL, eSOL, and iSOL (Molina-González, 2013), while others, like LIWC (Ramírez-Esparza et al., 2007) and SDAL, reflect various regional dialects of Spanish. While interest in Spanish sentiment analysis is growing, research resources remain scarce. Furthermore, Spanish is spoken across numerous countries and regions, where dialectal differences and vocabulary variations are pronounced. A single term may carry different meanings in different countries, or even between regions within the same country, complicating the development of unified sentiment analysis models. Regional slang and expressions further challenge accurate sentiment detection (Monsalve-Pulido, Parra & Aguilar, 2024). Spanish grammar and syntax add layers of complexity, particularly in double negations, flexible pronoun placement, and varied adjective ordering, which all impose unique demands on sentiment analysis models. For example, the placement of negation words can significantly alter sentence sentiment, and improper handling may result in incorrect classification (Angel, Negrón & Espinoza-Valdez, 2021).

Pre-trained large language models developed specifically for Spanish have begun to fill gaps left by models trained in other languages, providing more tailored solutions for sentiment analysis in Spanish. Among these, BETO (Cañete et al., 2020) and RoBERTuito (Pérez, J.M. et al., 2021) have shown marked advancements. BETO, the first BERT-based model tailored for Spanish, was trained on Spanish-language text from Wikipedia and OPUS, significantly outperforming the multilingual mBERT model across various tasks. RoBERTuito, a Spanish model based on the RoBERTa architecture and trained on over 500 million tweets, has demonstrated notable performance in sentiment analysis, emotion analysis, sarcasm detection, and hate speech detection. These models stand out due to their optimizations for Spanish grammar, vocabulary, and social media-specific structures, addressing gaps in pre-trained Spanish language models and providing precise solutions for Spanish-based tasks. This study, therefore, uses BETO and RoBERTuito for sentiment analysis tasks, capitalizing on their strengths in Spanish text processing.

## 2.3 Sentiment and Political Ideology in Spain

In Spain, political parties are distinctly positioned along the left-right political spectrum, reflecting their ideological stances and policy orientations. Podemos aligns with the far-left, emphasizing radical leftist policies such as opposition to austerity and support for social justice (Rama et al., 2023). The Spanish Socialist Workers' Party (PSOE) occupies a center-left position, advocating for social democracy and progressive reform (Khenkin, 2017). Ciudadanos, initially a center-right liberal party, has shifted its stance but still supports economic freedom and anti-corruption measures (Magre-Pont et al., 2021). On the right, the People's Party (PP) represents conservative values and a market economy (Khenkin, 2017), while Vox, positioned on the far-right, promotes anti-immigration, anti-globalization, and a strengthened sense of national sovereignty (Rama et al., 2023).

Recent studies suggest that emotional expression on social media, particularly Twitter, is closely related to these parties' political ideologies. Research by Garcia and Thelwall (2013) indicates that traditional parties like the PP and PSOE often express more neutral and positive sentiments, while emerging parties such as Vox are more likely to convey negative sentiments. This trend illustrates how political ideology can shape emotional communication strategies on social media platforms.

Since Spain's transition from a two-party to a multi-party system in 2015, the emergence of parties like Podemos and Ciudadanos has significantly impacted political stability and governance. This shift has redefined the political landscape, introducing new challenges and opportunities for how parties express emotions on social media (Khenkin, 2017). While recent studies have analyzed emotional expression by Spanish parties, they have primarily focused on qualitative analyses, particularly sentiment analysis for predicting election outcomes (Rodríguez-Ibáñez et al., 2021; Argandoña-Mamani et al., 2023). However, quantitative research on Spanish parties' sentiment and its alignment with political ideologies remains limited.

In this context, this study aims to explore Spanish political parties' sentiment on Twitter using pre-trained models, specifically applying BETO and RoBERTuito to analyze party tweets. By fine-tuning these models to enhance their performance for sentiment analysis tasks, the research seeks to determine whether they can accurately capture the sentiments of Spanish parties on social media and analyze the correlation between these sentiments and the parties' political positions. Through extensive fine-tuning, we aim to assess whether these sentiment analysis models can effectively reflect the emotional orientations of political parties, thus revealing connections between social media sentiment and ideological stances.

This approach not only provides valuable insights for political decision-making and public opinion monitoring but also offers a fresh perspective on strategic interactions on social media. By capturing dynamic changes in parties' emotional expression, this research enhances the quantitative methods used in Spanish political analysis, fostering a more objective and comprehensive understanding of how emotional sentiment aligns with political ideology.

Drawing parallels to the U.S. context, similar trends can be observed where emotional appeals in political communication have taken center stage, particularly during election cycles. American populist leaders have harnessed emotional rhetoric to galvanize support and deepen divides, demonstrating the pervasive power of emotions in contemporary politics. This comparison highlights the universal relevance of emotional strategies across political landscapes, underscoring the need for further research to explore these dynamics comprehensively.

**3. Research Questions**

Given the limited research leveraging pre-trained models to analyze sentiment in Spanish-language political tweets, this study aims to bridge that gap by utilizing the BETO and Robertuito models. Through this approach, we seek to address the following questions:

1. To what extent can large language models accurately capture sentiment variations among Spanish political parties with differing ideological orientations?
2. How do Spanish political parties with opposing ideologies differ in their use of emotional appeals, especially in times of crisis? What insights can these differences provide about their broader strategies for political communication and influence?

**4. The Present Study**

This study employs Spanish-specific pre-trained language models, BETO and RoBERTuito, to conduct sentiment analysis on tweets from major Spanish political parties, including PSOE, PP, Vox, Unidas Podemos, and Ciudadanos. Focusing on tweets from 2019 to 2024, the research aims to examine whether these models can accurately identify and quantify sentiment—positive, negative, or neutral—within each party's communications. By fine-tuning the models on Spanish-language datasets, the study seeks to enhance the models' performance in detecting sentiment nuances relevant to the political context. Ultimately, this research investigates the alignment between expressed sentiment and political ideology, providing insights into how Spanish political parties strategically use social media to engage with public sentiment and reflect their ideological positions.

**5. Methodology**

In this study, we fine-tuned the BETO and RoBERTuito models to enhance their performance for sentiment analysis in Spanish. This approach allows for an in-depth analysis of sentiment expression by Spanish political parties on Twitter and an exploration of its alignment with their political stances.

**4.1 Data Collection**

For the fine-tuning phase, we utilized a 2018 Spanish Twitter sentiment analysis dataset, which provides labeled data for sentiment classification and aligns with the objectives of the current sentiment analysis task. The dataset details are as Table 1:

**Table 1. Sentiment Distribution in the 2018 Spanish Twitter Dataset**

| Neutral | Positive | Negative |
|---|---|---|
| 111,334 | 11,004 | 9,489 |

In the inference phase, we gathered political tweets from Spanish users associated with PSOE, PP, VOX, Unidas Podemos, and Ciudadanos, covering the years 2019 to 2024. The tweet distribution by party is outlined in Table 2:

**Table 2. Distribution of Political Tweets by Party (2019-2023)**

| ciudadanos | podemos | pp | psoe | vox |
|---|---|---|---|---|
| 38592 | 48808 | 45480 | 61404 | 51505 |

The dataset consists of Twitter accounts associated with five major Spanish political parties: PP (Partido Popular), PSOE (Partido Socialista Obrero Español), Ciudadanos, VOX, and Podemos. For each party, the dataset includes two types of Twitter accounts:

1. Official Party Accounts: Each party's official Twitter handle (e.g., @ppoular for PP, @PSOE for PSOE) is included to capture the party's primary communication and official statements.

2. Key Representatives and Affiliates: The dataset also includes Twitter handles of notable representatives affiliated with each party, such as spokespersons, prominent lawyers (abogados), and activists who represent or support the party's views. These accounts are included to capture a broader range of political messaging, including personal perspectives, advocacy, and commentary that complement the official party line.

This selection provides a comprehensive view of each party's social media presence by incorporating both institutional and individual voices, allowing for an in-depth analysis of party communication, representative discourse, and affiliated commentary within the Spanish political landscape. Exemplary cases for the X accounts are displayed in Table 3.

**Table 3. Official and Key Twitter Accounts for Spanish Political Parties**

| party name | user id | party id |
|---|---|---|
| pp | @IdiazAyuso @TeoGarciaEgea @Aglezterol ... | @ppoular |
| psoe | @Hugo_Moran @patxilopez @JoseantonioJun ... | @PSOE |
| ciudadanos | @DaniPerezCalvo @InesArrimadas @GuillermoDiazCs ... | @CiudadanosCs |
| vox | @PabloSaezAM<br>@pedro_fhz<br>@Jorgebuxade<br>... | @vox_es |
| podemos | @ionebelarra<br>@IdoiaVR<br>@Irenirima<br>... | @PODEMOS |

## 4.2 Procedure

This study adheres to the methodological framework established by de Arriba et al. (2021), with a key distinction being our utilization of a significantly larger dataset for training, aimed at enhancing accuracy in sentiment analysis. Specifically, we fine-tuned our models on a labeled dataset comprising approximately 280,000 tweets, in contrast to the 8,223 tweets employed by de Arriba et al. (2021). The substantial size of our dataset is vital, as it augments the model's generalizability when applied to subsequent analyses of political tweets.

For finetuning this multi-class classification task, targeting three sentiment categories (negative, neutral, positive), the model was equipped with a fully connected layer mapping BERT's pooled output to three distinct sentiment classes. The optimization process was carried out using Stochastic Gradient Descent (SGD) with a learning rate set to $1*10^{-4}$, and the loss was calculated via CrossEntropyLoss, which is well-suited for categorical outcomes. The model was trained and evaluated using a batch size of 64 across a total of 20 epochs. Model checkpoints were saved based on improvements in test accuracy, initialized at a minimum threshold of 0.5. For accelerated training, we employed an NVIDIA RTX 4090, with Stochastic Gradient Descent (SGD) as the optimization algorithm. Throughout the training process, we closely monitored convergence to mitigate the risk of overfitting, ensuring that the models effectively learned from the training data. The detailed workflow is illustrated in Figure 1. To assess model performance, we utilized key metrics including training accuracy, test accuracy, and loss.

**Figure 1. Workflow Diagram for Sentiment Analysis Process**

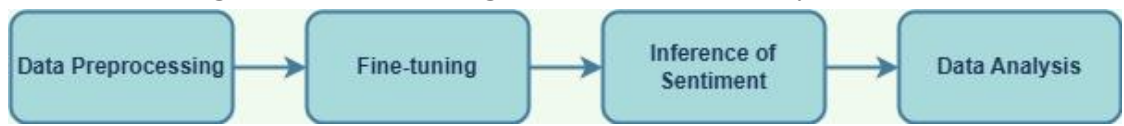

## 4.3 Data Preprocessing

We began preprocessing the data by performing basic cleaning on the Spanish-language corpus. Given the dataset's Twitter origins(see Figure 2), it included considerable internet jargon and expressions such as "@" (for mentions) and "#" (for hashtags). These elements, often irrelevant to sentiment analysis, were removed. The preprocessing steps included:

1. Removal of URLs and Emojis: URLs were deleted, as they generally lack meaningful sentiment information and could impair the accuracy of the analysis. Similarly, emojis were removed due to their ambiguous sentiment implications.

2. Mentions: User mentions were eliminated to reduce noise and ensure user privacy.

3. Stopword Filtering: Common Spanish stopwords (e.g., "uno," "el," "que") were filtered out to enhance model performance by reducing irrelevant noise.

4. Lemmatization: We reduced various conjugated forms of Spanish verbs to their root forms, thereby decreasing the feature space and improving model accuracy.

Finally, the cleaned text was processed using a model-specific tokenizer.

**Figure 2. Sample Spanish Tweet**

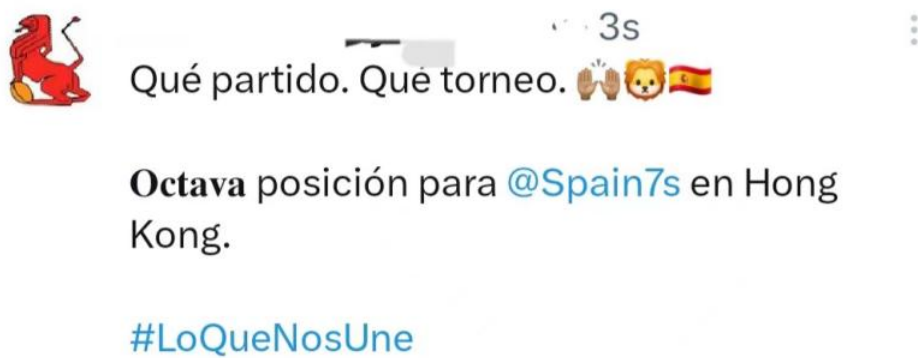

**4.4 Model Selection**

Research by Pérez, J. M. et al. (2021) highlights the superior performance of the pre-trained models BETO and RoBERTuito in Spanish-language tasks, thanks to their specialized training and optimization, particularly for sentiment analysis. As a result, we chose these models for sentiment analysis of Spanish tweets to evaluate their effectiveness in identifying sentiment polarity within social media texts and to compare their performance on this task.

BETO is the first BERT model pre-trained exclusively for Spanish, using a large corpus of Spanish-language texts from sources like Wikipedia and the OPUS project. The data volume used is comparable to the original BERT's corpus. BETO also incorporates successful training techniques from previous models, such as dynamic masking and Whole-Word Masking (WWM). It has shown outstanding performance across various tasks, consistently outperforming mBERT, and achieving state-of-the-art results on standard Spanish datasets (e.g., POS and MLDoc).

RoBERTuito, another pre-trained language model, is optimized for Spanish user-generated content. Built on the RoBERTa architecture, RoBERTuito was trained on a corpus of over 500 million tweets, adopting successful hyperparameters from earlier models and using a large batch size. In classification tasks on Spanish Twitter data—sentiment analysis, emotion analysis, sarcasm detection, and hate speech detection—RoBERTuito performs significantly better than other pre-trained Spanish language models, particularly excelling in hate speech detection and sentiment analysis.

While both BETO and RoBERTuito are designed for Spanish, their architectural foundations and training methodologies differ. BETO's architecture is closely aligned with BERT, focusing on masked language modeling and the ability to understand sentence relationships through the use of attention mechanisms. In contrast, RoBERTuito adopts the RoBERTa approach, which emphasizes training with larger batches and more extensive data, optimizing the model's performance for dynamic and informal user-generated content prevalent on social media.

BETO and RoBERTuito's extensive training on Spanish-language text and optimizations for Spanish-specific characteristics give them a competitive edge for Spanish-language tasks. These models account for Spanish grammar, vocabulary patterns, and social media text structures, enhancing their ability to understand and process Spanish text. General pre-trained models may not adapt as well to Spanish's specific nuances, so using dedicated Spanish models like BETO and RoBERTuito is likely to yield superior results.

In selecting models for this study, we employed BETO and RoBERTuito, two pre-trained models optimized for Spanish-language text, to cross-validate our findings. Using both models allowed us to confirm that our sentiment classification results were not model-dependent, thereby enhancing the robustness and reliability of our analysis. Each model captures different linguistic patterns due to variations in pre-training data and architecture, providing complementary perspectives on Spanish sentiment. By observing consistent results across both BETO and RoBERTuito, we ensured that the sentiment trends identified in our analysis were authentic and free from model-specific bias, strengthening the validity and reproducibility of our findings.

### 4.5 Experimental Environment
The configuration details of the experimental environment used in this study are provided in Table 4.

Table 4. System Configuration for Model Training

| Component | Specification |
|---|---|
| Operating System | Ubuntu 20.04 |
| Programming Language | Python 3.8 |
| Training Framework | Pytorch 1.11.0 + cuda 11.3 |
| GPU | RTX 4090(24GB) |
| CPU | 16 vCPU Intel(R) Xeon(R) Gold 6430 |

### 5. Results

**Finetuning results**

In this study, our fine-tuning approach achieved superior performance metrics relative to previous benchmarks, specifically outperforming de Arriba et al. (2021) in sentiment classification accuracy. Using the BETO model with similar hyperparameters (learning rate of $1*10^{-4}$) but an extended training dataset, we achieved a test accuracy of 0.9024, along with a macro-averaged F1 score of 0.9024, a precision of 0.9022, and a recall of 0.9024. Additionally, when fine-tuning the Robertuito model under similar conditions, we obtained a test accuracy of 0.9011, an F1 score of 0.9008, a precision of 0.8901, and a recall of 0.9011.

For a robust evaluation, we allocated 40% of the dataset for testing, thus enhancing the reliability of the performance metrics and reducing potential overfitting concerns. Both models demonstrated significant improvements over the metrics reported in de Arriba et al. (2021) across all evaluation measures, as shown in **Table 5**.

The training progression, depicted in Figures 3 and 4, illustrates the performance of both models over the 20 epochs. Notably, both training and test accuracies exhibit rapid increases during the initial epochs, indicating effective early learning and a well-calibrated learning rate. After several epochs, the test accuracy stabilizes around 90%, suggesting strong generalization capabilities for each model. Training accuracy continues to rise gradually, eventually aligning closely with test accuracy, which alleviates concerns regarding overfitting.

The loss curves in both figures demonstrate a consistent decline, with significant drops in the early epochs that gradually taper off as training continues. This steady reduction in loss signifies smooth convergence and affirms the stability of the training process. Despite minor variations in the final accuracy and loss values between the two models, both configurations showcase robust performance, achieving high accuracy alongside minimized loss, indicating that the models are well-optimized for the sentiment analysis task on the specified dataset.

The accuracy, F1 score, precision and call are demonstrated as follows:

**Table 5. Model Performance Comparison with Models in de Arriba et al,(2021)**

|  | Model | Accuracy | F1 Score | precision | recall |
|---|---|---|---|---|---|
| de Arriba et al.(2021) | 1 | 0.644324 | 0.612514 | 0.59436 | 0.644324 |
|  | 2 | 0.652778 | 0.622223 | 0.600479 | 0.652778 |
|  | 3 | 0.641908 | 0.626723 | 0.625913 | 0.641908 |
| this study | BETO | 0.9024 | 0.9024 | 0.9022 | 0.9024 |
|  | Robertuito | 0.9011 | 0.9008 | 0.8901 | 0.9011 |

**Sentiment analysis results**

After fine-tuning, we saved the model parameters and proceeded to perform inference on the Spanish political tweet dataset utilizing the trained models. This dataset includes timestamps and the political affiliations of users, with tweet content serving as the primary input for sentiment predictions. Descriptive results using BETO and RoBERTuito are demonstrated in

Table 6 and Table 7.

Table 6. Sentiment Analysis Results Using BETO

| Partido | Negative | Positive | Neutral |
|---|---|---|---|
| ciudadanos | 3304 | 5610 | 29591 |
| podemos | 3670 | 5524 | 39001 |
| pp | 3061 | 6073 | 36165 |
| psoe | 3567 | 6617 | 50968 |
| vox | 5180 | 5768 | 39965 |

Table 7. Sentiment Analysis Results Using RoBERTuito

| Partido | Negative | Positive | Neutral |
|---|---|---|---|
| ciudadanos | 3805 | 6350 | 28350 |
| podemos | 4431 | 6685 | 37079 |
| pp | 3801 | 7070 | 34428 |
| psoe | 4231 | 7524 | 49397 |
| vox | 6119 | 6703 | 38091 |

**Chi-square Test Results**

A Chi-Square test was performed to determine whether the observed differences in sentiment distributions between the BETO and RoBERTuito models were statistically significant for each political party(as in Table 8). The results for all parties indicated highly significant differences ($p < 0.0001$).

Table 8. Chi-Square Test Results for Sentiment Distribution Across Political Parties

| Political Party | Chi-Square ($\chi^2$) | p-value |
|---|---|---|
| Ciudadanos | 107.67 | < 0.0001 |
| Podemos | 230.45 | < 0.0001 |
| PP | 198.17 | < 0.0001 |
| PSOE | 139.31 | < 0.0001 |
| Vox | 193.13 | < 0.0001 |

These results indicate statistically significant differences in sentiment distributions between the BETO and RoBERTuito models for all political parties, suggesting that the models interpret and classify sentiments differently across the political spectrum.

**Correlation Analysis**

To further evaluate the alignment of sentiment trends across models, we calculated Pearson correlation coefficients for each sentiment category (Negative, Positive, and Neutral) across both models. The results show high correlation values(see Table 9), indicating a strong consistency in overall sentiment trends despite the significant differences identified by the Chi-Square test

Table 9. Correlation Analysis of Sentiment Classifications Between BETO and RoBERTuito

| Sentiment | Correlation | p-value |
|---|---|---|
| Negative | 0.99 | 0.00052 |
| Positive | 0.94 | 0.01685 |
| Neutral | 1 | 0.00002 |

The strong positive correlations across all sentiment categories suggest that while BETO and RoBERTuito may differ in exact proportions, both models capture similar sentiment trends for each political party.

**Confidence Intervals for Sentiment Proportions**

Finally, we calculated 95% confidence intervals for Negative, Positive, and Neutral sentiments within each party for both BETO and RoBERTuito(in Table 10). These confidence intervals provide insight into the range within which the true sentiment proportions likely fall.

Table 10: Sentiment Proportion Ranges by Political Party for BETO and RoBERTuito

| Model | Partido | Negative (%) | Positive (%) | Neutral (%) |
|---|---|---|---|---|
| **BETO** | Ciudadanos | 8.30 - 8.86 | 14.22 - 14.92 | 76.43 - 77.27 |
| | Podemos | 7.38 - 7.85 | 11.18 - 11.75 | 80.57 - 81.27 |
| | PP | 6.53 - 6.99 | 13.09 - 13.72 | 79.47 - 80.21 |
| | PSOE | 5.65 - 6.02 | 10.57 - 11.07 | 83.05 - 83.64 |
| | Vox | 9.91 - 10.44 | 11.05 - 11.60 | 78.14 - 78.85 |
| **RoBERTuito** | Ciudadanos | 9.58 - 10.18 | 16.12 - 16.86 | 73.19 - 74.07 |
| | Podemos | 8.94 - 9.45 | 13.56 - 14.18 | 76.56 - 77.31 |
| | PP | 8.14 - 8.65 | 15.27 - 15.94 | 75.61 - 76.39 |
| | PSOE | 6.72 - 7.12 | 12.04 - 12.56 | 80.47 - 81.09 |
| | Vox | 11.74 - 12.30 | 12.87 - 13.46 | 74.44 - 75.19 |

The confidence intervals reveal differences in sentiment proportions between the two models. For example, the BETO model consistently shows higher Neutral sentiment proportions and lower Negative sentiment proportions than the RoBERTuito model across most parties. This suggests that BETO might be more conservative in assigning Negative sentiment, while RoBERTuito identifies a relatively higher proportion of tweets as Negative, particularly for parties such as Vox and Ciudadanos.

These tables allow for several key observations. Although neutral sentiment is predominant across party tweets, there are noticeable differences in positive and negative sentiment among parties. PSOE tends to publish more positive tweets, while Vox displays a greater share of negative sentiment.

**Figure 3: Sentiment Distribution by Political Party According to BETO**

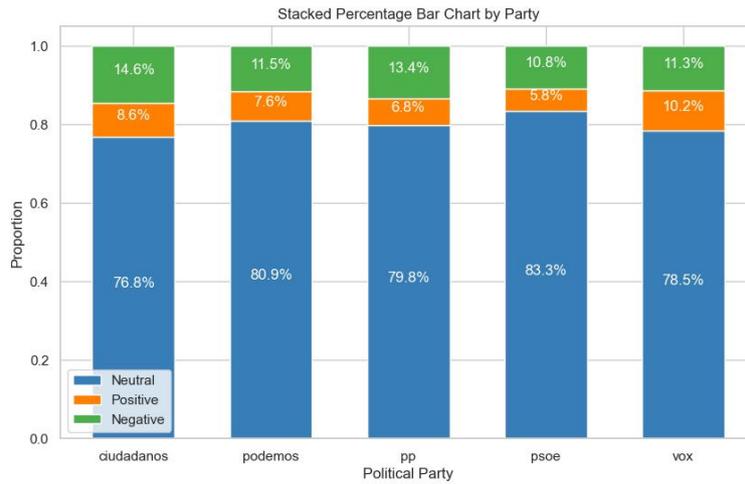

**Figure 4: Sentiment Distribution by Political Party According to RoBERTuito Model**

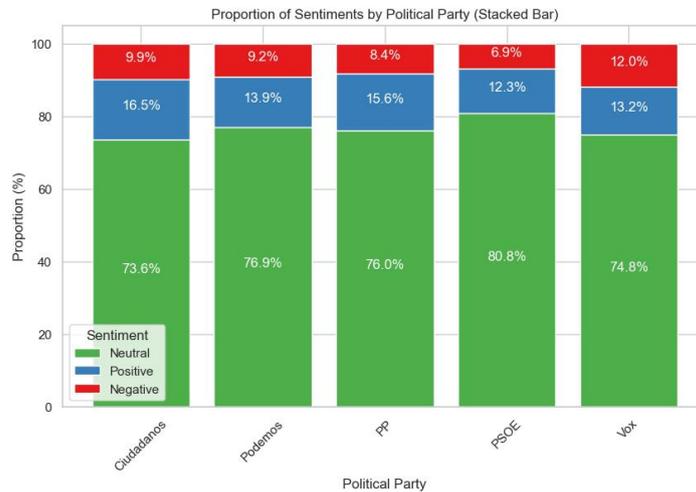

Figures 5 illustrate the range of sentiment trends among Spanish political parties on Twitter from 2019 to 2024. PSOE's sentiment is relatively stable, with a slight rise in positive sentiment and consistent neutral sentiment. By contrast, Ciudadanos shows more variation, with a slight overall decline in positive and neutral sentiment. Vox, however, reveals a steady increase in negative sentiment and a corresponding decrease in positive sentiment. PP's sentiment trend shows moderate growth in positive sentiment while keeping negative sentiment stable. Podemos has a relatively stable trend, with only a slight decline in neutral sentiment. From a comparative standpoint, PSOE and Ciudadanos lean more toward positive tweets, whereas Vox and PP tend more toward neutral or negative tweets. Between 2019 and 2020, Ciudadanos displayed a higher proportion of positive sentiment, which later decreased, while Vox saw an increase in negative sentiment over this period. Between 2021 and 2023, PSOE's positive sentiment grew slightly, while Vox's negative sentiment continued to rise.

**Figure 5: Yearly Trends in Positive and Negative Sentiment Proportions by Political**

**Party (2019-2024)**

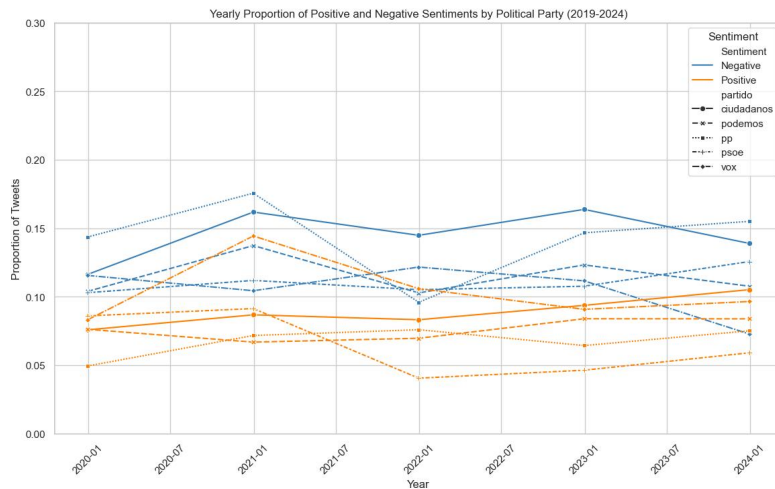

Calculating the "neutral + positive" sentiment proportion in each party's tweets (Figure 6) shows that PSOE holds the highest proportion at 94.17%, followed by PP at 93.24%, Podemos at 92.39%, Ciudadanos at 91.42%, and Vox the lowest at 89.83%. Negative sentiment proportions are 5.83% for PSOE, 6.76% for PP, 7.61% for Podemos, 8.58% for Ciudadanos, and highest for Vox at 10.17%, as shown in Figure 16. Chi-square test results show significant differences between sentiment categories ($\chi^2$=838.06, df=4, p-value < 2.2e-16), confirming that sentiment expression varies significantly among the different political parties on Twitter.

**Figure 6: Proportion of Neutral + Positive and Negative Sentiment by Political Party**

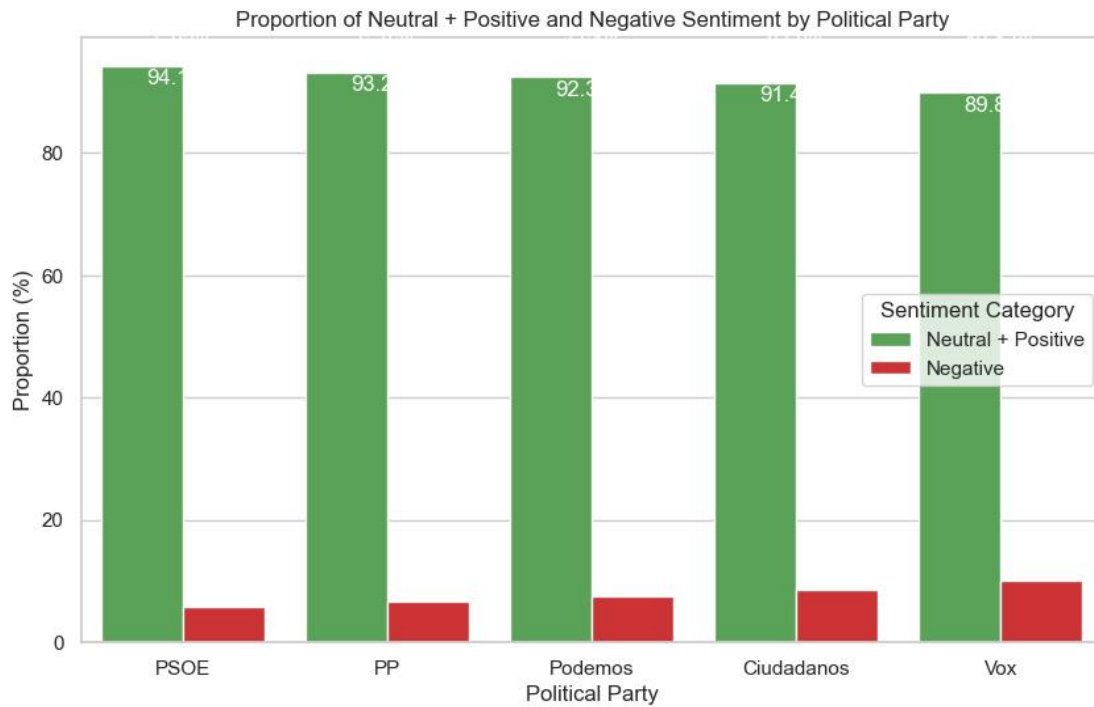

**6. Discussion**

This study fine-tuned Spanish pre-trained BERT and RoBERTa models on a labeled Spanish Twitter sentiment analysis dataset, then used these models to analyze the sentiment tendencies in Spanish political parties' tweets from 2019 to 2024. The results reveal a diverse range of sentiment tendencies among Spanish political parties, with some maintaining relatively stable sentiment and others showing considerable fluctuation. We found a strong association between the sentiment tendencies of parties and their political stances. For example, PSOE and Ciudadanos generally use more neutral and positive sentiment in their tweets, while Vox and PP lean toward neutral and negative sentiment, reflecting each party's political stance. These findings offer valuable insights into the sentiment expression of Spanish political parties on social media and its relationship to their political positions. Through a quantitative analysis of sentiment in tweets by major Spanish parties, this study highlights connections between sentiment tendencies, political stances, and social media strategies, showcasing significant advantages and advancements over previous studies that relied primarily on qualitative analysis.

**6.1 Model Performance**
In this study, using pre-trained models has significantly enhanced the accuracy and efficiency of sentiment analysis on Spanish political tweets. BETO and RoBERTuito, models specifically pre-trained on extensive Spanish-language corpora, excel at capturing the intricacies of Spanish syntax, lexical nuances, and colloquial expressions typical in tweets. By leveraging a larger training dataset, this study optimized model performance, yielding improved accuracy and task-specific robustness. Unlike qualitative approaches that depend on subjective interpretations, this data-driven approach ensures consistency and objectivity across a vast dataset, reducing human bias and bolstering the validity and reproducibility of findings. The increased training data was instrumental in refining model predictions, underscoring the power of pre-trained models in scaling sentiment analysis with reliable, high-performance results(de Arriba et al., 2021).

In addition, this study employs two models applicable for the Spanish language, though they exhibit distinct approaches in sentiment classification, each showing unique tendencies that influence the overall sentiment proportions assigned to each political party. BETO demonstrates a tendency toward higher Neutral sentiment classifications across all parties, as shown by its consistently larger confidence intervals for the Neutral category compared to RoBERTuito. This conservative approach in assigning sentiment suggests that BETO may require stronger linguistic cues to identify tweets as either Positive or Negative. For instance, BETO classified between 76.43% and 77.27% of Ciudadanos-related tweets as Neutral, compared to RoBERTuito's 73.19% to 74.07%.

In contrast, RoBERTuito tends to classify more tweets as either Positive or Negative, which reflects a sensitivity to nuanced sentiment expressions that BETO might categorize as Neutral. For example, RoBERTuito shows a broader Positive sentiment range for Ciudadanos (16.12% to 16.86%) and PP (15.27% to 15.94%) than BETO. This pattern suggests that RoBERTuito is more responsive to subtle language cues that BETO does not capture as readily. Additionally, RoBERTuito assigns a consistently higher proportion of Negative sentiment for certain parties, such as Vox, where it classifies 11.74% to 12.30% of tweets as Negative compared to BETO's 9.91% to 10.44%. These distinctions underscore the impact of model selection, as each model's design influences the perception of sentiment polarity and strength within the data.

**6.2 Sentiment Proportion in the Spanish Political Parties**
This study quantitatively validates prior qualitative hypotheses, illuminating the connection between political parties' sentiment tendencies and their social media strategies in Spain. By employing pre-trained sentiment analysis models, this research offers a data-driven perspective on sentiment expression across Spanish political parties and supports the idea that political ideology strongly influences sentiment strategy. In doing so, this study builds on established literature by extending our understanding of political communication through quantitative methods, specifically for Spanish-language data, which has historically been underrepresented in sentiment research.

Both BETO and RoBERTuito models reveal a consistent sentiment pattern across parties, with Neutral sentiment dominating, particularly in PSOE's discourse, as shown by BETO's 83.64% and RoBERTuito's 81.09% classifications. This supports findings by Garcia and Thelwall (2013), which suggested that traditional parties like PSOE and PP often rely on more moderate language to maintain a broader appeal. By contrast, the study's lower Neutral sentiment proportions for parties like Vox and Ciudadanos align with the hypothesis that opposition or far-right parties utilize more emotionally charged, polarized language to differentiate their messages. This is consistent with observations by Kluknavská and Macková (2024) and Fourie et al. (2022) who noted that opposition parties often employ emotional rhetoric—anger, fear, and uncertainty—to resonate with disenfranchised or frustrated voters.

The sentiment trends observed here illustrate the strategic use of emotional appeals. PSOE's high Neutral and Positive sentiment proportions reflect its approach as a center-left party, appealing to progressive ideals through resilience, unity, and hope. This strategic moderation aligns with studies showing that mainstream parties often avoid divisive rhetoric to foster social cohesion (Peitz et al., 2021). In contrast, Vox's significant Negative sentiment proportion underscores its far-right positioning, where sentiment is used to foster a sense of urgency and threat. This approach, noted by Rama et al. (2023), illustrates how populist parties leverage emotional appeals to galvanize support, particularly by emphasizing themes of national sovereignty and identity.

In addition to examining sentiment trends across parties, this study highlights the broader implications of political messaging on public discourse, especially in contexts like Spain's multi-party system where ideological diversity has reshaped political engagement. The strategic variations in emotional appeal observed between traditional and emerging parties underscore the literature's insights on how emotional dynamics can shape public opinion and potentially reinforce ideological divides (Druckman & McDermott, 2008). In particular, these findings confirm the literature's assertions that far-right and opposition parties may utilize heightened emotional appeals to position themselves as defenders of national integrity, contrasting sharply with the more unifying narratives of centrist parties.

Through this study, we demonstrates the effectiveness of quantitative sentiment analysis in validating hypotheses traditionally explored through qualitative methods. By using advanced models fine-tuned for Spanish, this study bridges the gap between sentiment analysis research predominantly focused on English and other languages. The successful adaptation of BETO and RoBERTuito in this context reinforces the potential for quantitative models to complement qualitative analyses, as shown in this study's ability to reflect the political and emotional nuances that define Spain's current political landscape.

### 6.3 Future Implications

This study sheds light on the use of social media as a platform for political discourse, with particular relevance to Spanish-speaking regions, where research on political sentiment in social media remains relatively scarce compared to English-language studies. By analyzing Spanish political parties' social media strategies, this work enhances our understanding of the nuances of political communication in Spain and sets a foundation for similar studies in other Spanish-speaking countries. Expanding sentiment analysis in Spanish is essential, as it enriches the global perspective on political sentiment, enabling meaningful comparisons across different linguistic and cultural contexts and addressing a significant gap in non-English political sentiment research.

From a methodological perspective, future studies could address some limitations observed in this analysis, particularly the dominance of Neutral sentiment classifications. This trend suggests the need for more nuanced categories (e.g., slight positive or slight negative) that could capture subtle emotional undertones in political messaging. Introducing such granularity would help distinguish truly neutral content from statements that convey mild positivity or negativity, thereby providing a more refined understanding of public sentiment.

Additionally, while this study employed pre-trained models like BETO and RoBERTuito, it is important to consider the potential biases inherent in these models(Liang et al., 2021; Ferrara, 2023), especially when analyzing nuanced political sentiments such as sarcasm or mixed emotions. These pre-trained models, though fine-tuned for Spanish, may still carry biases from their original training data, which could influence sentiment classification in complex or ambiguous cases. Addressing these biases in future research could enhance the reliability of sentiment analysis in Spanish, making it more robust and reflective of the true public mood.

Broadening the scope to include cross-linguistic comparisons of sentiment trends would also enhance the relevance of these findings. Conducting comparative studies in other languages, particularly with English, could reveal whether similar sentiment structures and political communication patterns appear across diverse political and cultural landscapes. Such comparisons could not only contextualize Spanish-language findings but also contribute to a more holistic understanding of sentiment expression in global political narratives.

Lastly, the extensive use of hashtags on Twitter represents a valuable area for future exploration. In this study, hashtags were removed for simplicity, but they often serve as thematic markers and rallying points for campaigns (e.g., interpreting "#Spain" as "Spain" could connect tweets to broader topics of national identity). Future research could incorporate hashtag analysis(eg. Alfina et al., 2017; Nazir et al., 2019), either by transforming them into keywords or analyzing them as standalone topics, to uncover the broader discourse surrounding political sentiment. This approach could reveal additional layers of meaning within public sentiment, providing insights into popular themes and narratives that resonate with audiences.

In addressing these areas—model biases, sentiment granularity, cross-linguistic comparisons, and hashtag analysis—future research can deepen our understanding of political discourse on social media, especially in underexplored languages like Spanish. Such work has practical implications for political analysts, policymakers, and social media platforms striving to engage and comprehend diverse audiences within increasingly polarized digital environments.

## 7. Conclusion

This experiment utilized Spanish pre-trained models based on BERT and RoBERTa encoder architectures, which offer strong performance despite a relatively modest parameter size. However, with the emergence of large language models like GPT-4 and Claude2, models with hundreds of billions of parameters are anticipated to possess broader general knowledge capabilities. In the future, as computational resources expand, we could extend this study's tasks to larger models for further experimentation.

In this study, sentiment classification was set as a three-class problem (positive, negative, neutral), balancing classification effectiveness with the depth of insights extracted from the data. As a critical component of public opinion analysis on social media, employing more nuanced and fine-grained sentiment categories could uncover deeper insights, albeit possibly at some cost to model performance and accuracy.

Moreover, this study used only tweet content as input for sentiment analysis. However, social media platforms provide a wealth of additional information, such as user profiles, historical data, and social networks. By incorporating these richer features, models could better capture subtle sentiment nuances, enabling a more comprehensive analysis and, when combined with party political stances, offering deeper insights.